\documentclass[11pt]{article}

\usepackage[final]{acl}

\usepackage{times}
\usepackage{latexsym}
\usepackage{amsmath}
\usepackage{amssymb}
\usepackage{multirow}
\usepackage{xcolor}
\usepackage{colortbl}
\usepackage{float} 
\usepackage{pifont}  
\usepackage[T1]{fontenc}

\usepackage[utf8]{inputenc}

\usepackage{microtype}

\usepackage{graphicx}

\usepackage{booktabs}

\usepackage{pifont}

\usepackage{enumitem}

\usepackage{float}

\definecolor{bestgreen}{RGB}{34, 139, 34}
\definecolor{warnorange}{RGB}{255, 140, 0}
\definecolor{badred}{RGB}{220, 20, 60}
\definecolor{lightgreen}{RGB}{230, 255, 230}
\definecolor{lightgray}{RGB}{245, 245, 245}
\definecolor{gtblue}{RGB}{30, 100, 180}

\title{StoryTR: Narrative-Centric Video Temporal Retrieval \\with Theory of Mind Reasoning}

\author{
  \textbf{Xuanyue Zhong}\textsuperscript{1}\thanks{~~Equal contribution.}\thanks{~~Work done during research internship at Skywork AI.}, 
  \textbf{Yuqiang Xie}\textsuperscript{2}\footnotemark[1], 
  \textbf{Guanqun Bi}\textsuperscript{3}, 
  \textbf{Jiangping Yang}\textsuperscript{2}, 
  \textbf{Guibin Chen}\textsuperscript{2}\thanks{~~Corresponding author.} \\
  \textsuperscript{1}Department of Computer Science and Engineering, \\ The Chinese University of Hong Kong, Hong Kong \\
  \textsuperscript{2}SkyReels, Skywork AI \\
  \textsuperscript{3}Independent Researcher \\
  \texttt{1155242700@link.cuhk.edu.hk, indexfziq@gmail.com}
}

\begin{document}
\maketitle
\begin{abstract}
Current video moment retrieval excels at action-centric tasks but struggles with narrative content. Models can see \textit{what is happening} but fail to reason \textit{why it matters}. This semantic gap stems from the lack of \textbf{Theory of Mind (ToM)}: the cognitive ability to infer implicit intentions, mental states, and narrative causality from surface-level observations. We introduce \textbf{StoryTR}, the first video moment retrieval benchmark requiring ToM reasoning, comprising 8.1k samples from narrative short-form videos (shorts/reels). These videos present an ideal testbed. Their high information density encodes meaning through subtle multimodal cues. For instance, a glance paired with a sigh carries entirely different semantics than the glance alone. Yet multimodal perception alone is insufficient; ToM is required to decode that a character ``smiling'' may actually be ``concealing hostility.'' To teach models this reasoning capability, we propose an \textbf{Agentic Data Pipeline} that generates training data with explicit three-tier ToM chains (intent decoding, narrative reasoning, boundary localization). Experiments reveal the severity of the reasoning gap: Gemini-3.0-Pro achieves only 0.53 Avg IoU on StoryTR. However, our 7B \textbf{Shorts-Moment} model, trained on ToM-guided data, improves +15.1\% relative IoU over baselines, demonstrating that \textit{narrative reasoning capability matters more than parameter scale}.
\end{abstract}

\section{Introduction}


Video moment retrieval (VMR) has achieved remarkable success on action-centric benchmarks~\citep{qvhighlights,gao2017charadeSTA,hendricks2017didemo}, where queries like ``find the person running'' can be resolved through direct visual pattern matching.
However, when confronted with \textit{narrative-centric} content, current multimodal large language models (MLLMs) exhibit a critical \textbf{semantic gap}: they excel at recognizing \textit{explicit} visual patterns but struggle to infer \textit{implicit} narrative intents.
In essence, \textit{current models are skilled at seeing ``what is happening'' explicitly, but fail to reason ``why it matters'' implicitly}.
This gap becomes pronounced when visual surfaces contradict internal mental states. A character may be smiling while plotting revenge, or appear calm while harboring deep grief.


Narrative short-form videos (shorts/reels) serve as the ideal testbed for this challenge.
Unlike simple video clips, shorts are \textit{high-density streams of social signals}. They compress complete story arcs into 1-3 minutes through dialogue reversals, sudden BGM shifts, and close-up micro-expressions.
This extreme information density demands \textbf{native multimodal perception}: a protagonist's glance (visual) paired with a sigh (audio) carries entirely different semantics than the glance alone; a visually calm scene overlaid with dissonant music signals hidden tension.
Yet multimodal perception alone is insufficient.
Consider the query ``Find the moment the protagonist realizes the betrayal.''
The visual surface may show only silence and a lingering gaze, while the audio reveals a subtle tremor in the voice.
Multimodal models can \textit{perceive} these cues but cannot \textit{interpret} them: they see ``a person pausing'' but fail to infer ``realization of betrayal.''
Bridging this gap requires \textbf{Theory of Mind (ToM)}~\citep{premack1978theory}. ToM is the cognitive ability to attribute mental states, beliefs, and intentions to others. It transforms scattered multimodal cues into coherent narrative understanding.


We hypothesize that ToM reasoning, while absent in current models, is \textit{learnable through data}.
Traditional VMR annotations provide mere timestamps without reasoning traces. They cannot teach models \textit{why} a moment matters narratively.
To address this, we propose an \textbf{Agentic Data Pipeline} that generates training data with explicit reasoning chains.
The pipeline employs a \textit{Clipper Agent} for fine-grained multimodal perception (capturing subtle visual and audio cues), and a \textit{Self-QA Agent} that produces answers guided by \textbf{three-tier ToM reasoning}. The three tiers are: (1) \textit{Intent Decoding}, which infers character goals and mental states; (2) \textit{Narrative Reasoning}, which traces causal chains and plot significance; and (3) \textit{Boundary Localization}, which grounds abstract understanding to precise temporal segments.
This approach distills ToM capabilities from advanced models (Gemini-3.0-Pro) into structured training data, enabling smaller models to acquire narrative reasoning through explicit chain-of-thought supervision.


Our experiments quantify the severity of this cognitive gap: even Gemini-3.0-Pro achieves only 0.53 Avg IoU on narrative retrieval, while action-optimized models like Qwen3-Omni collapse to 0.07.
Crucially, our 7B Shorts-Moment model, trained on ToM-guided data, achieves +15.1\% relative IoU improvement, outperforming larger models and demonstrating that \textit{reasoning capability matters more than parameter scale}.

We make three contributions:
\begin{itemize}[leftmargin=*,itemsep=0pt,topsep=2pt]
\item \textbf{StoryTR Benchmark}: The first VMR benchmark requiring Theory of Mind reasoning, with 8.1k samples designed to test intent decoding, causal reasoning, and narrative understanding.
\item \textbf{ToM-Guided Data Paradigm}: A principled approach demonstrating that explicit reasoning chains can transfer cognitive capabilities from large to small models, addressing the ``invisible intent'' problem in video understanding.
\item \textbf{Empirical Validation}: Evidence that narrative reasoning is learnable. Our 7B model surpasses 30B+ baselines, proving that cognitive depth outweighs computational scale.
\end{itemize}

\section{Related Work}


\subsection{Video Moment Retrieval: Success and Limitations}

Video Moment Retrieval (VMR) aims to localize temporal segments corresponding to natural language queries. Early approaches framed this as cross-modal alignment~\citep{gao2017charadeSTA,hendricks2017didemo}, with benchmarks like QVHighlights~\citep{qvhighlights} and TVR~\citep{lei2020tvr} driving progress on action-centric content.

The advent of multimodal large language models (MLLMs) has significantly advanced VMR capabilities. Models such as InternVideo2~\citep{wang2025internvideo2}, Qwen-VL~\citep{bai2025qwen25vl,qwen3omni2025}, and InternVL3~\citep{zhu2025internvl3} demonstrate strong cross-modal alignment, while native multimodal architectures like Gemini~\citep{googledeepmind2025gemini3} and ARC-Hunyuan~\citep{ge2025archunyuan} process video, audio, and text as unified tokens.

\textbf{The Perception-Cognition Gap.} Despite these advances, current VMR systems share a fundamental limitation: they excel at \textit{perceptual} tasks (detecting ``when someone jumps'') but struggle with \textit{cognitive} tasks (understanding ``why the jump matters narratively''). This gap manifests in two dimensions. First, \textit{temporal duration}: benchmarks like Video-MME~\citep{fu2024videomme} and LongVideoBench~\citep{wu2024longvideobench} reveal that reasoning degrades with video length. Second, and more fundamentally, \textit{semantic depth}: existing datasets lack queries requiring inference about character intent, emotional dynamics, or narrative causality. Our work addresses this second dimension by moving VMR from pattern matching to narrative reasoning.

\subsection{Narrative Understanding and Theory of Mind}

\textbf{From Perception to Cognition.} Existing narrative benchmarks like MovieNet~\citep{huang2020movienet} and ShotBench~\citep{liu2025shotbench} evaluate \textit{what} happens (shot boundaries, scene transitions) rather than \textit{why} moments matter (character mental states, plot significance). Models can segment videos but cannot explain narrative function. Narrative short-form videos amplify this challenge by compressing story arcs into 1-3 minutes as \textit{high-density signal streams}. Unlike long-form content where narrative cues are distributed across hours, every second in shorts carries critical story weight. A model may detect ``a woman smiling'' but fail to infer ``she is concealing hostility,'' which requires reasoning about mental states invisible to pure perception.

\textbf{ToM as Solution.} Theory of Mind (ToM)~\citep{premack1978theory,wellman2001meta}, the capacity to attribute mental states and intentions to others, has proven effective for dialogue understanding~\citep{sap2022neural} and text-based story comprehension~\citep{rashkin2018modeling,kosinski2023theory,xie2022comma}. ToM enables inferring what characters believe, what they intend, and why their actions matter within the narrative structure. Our work bridges this gap by introducing ToM reasoning to VMR, demonstrating that the ``invisible intent'' problem can be addressed through explicit cognitive reasoning rather than improved perception alone.

\section{The StoryTR Benchmark}


We construct StoryTR (Story-centric Temporal Retrieval), a benchmark explicitly designed to test \textbf{Theory of Mind reasoning} in video moment retrieval. Unlike action-centric benchmarks that evaluate perceptual capabilities (``when does someone run?''), StoryTR evaluates cognitive capabilities (``when does the protagonist realize the betrayal?''). This distinction is fundamental: the former requires pattern matching, the latter requires reasoning about mental states.

\subsection{Why Short Dramas?}

\textbf{Design Rationale.} We select narrative short-form videos (shorts/reels) not for novelty, but because they constitute an ideal testbed for the perception-cognition gap. Short dramas are \textit{high-density streams of social signals}. They compress complete story arcs into 1-3 minutes through dialogue reversals, BGM shifts, and micro-expressions. This density creates a rigorous evaluation environment: every second carries narrative weight, and success requires understanding \textit{why} moments matter, not just \textit{what} happens.

\textbf{Data Source.} We collect videos featuring character development, emotional climaxes, and plot twists comparable to feature-length content. This compressed storytelling amplifies the ToM challenge. A protagonist's subtle glance may signal betrayal, jealousy, or reconciliation depending on narrative context. This information is invisible to perception-only models.

\textbf{Scale and Quality.} StoryTR comprises 8,141 samples (7,330 training, 811 testing). We prioritize \textit{annotation depth} over scale: each query-answer pair is designed to require genuine narrative reasoning, with 811 test samples human-calibrated for rigorous evaluation (see Appendix~\ref{sec:appendix_annotation}).

\textbf{Data Consent and Usage Rights.} All videos in StoryTR are sourced from publicly available short-form video platforms where content is released under commercial licenses permitting research use. We obtained explicit permission from content distributors for academic research purposes. Our dataset contains only metadata annotations (queries, timestamps, reasoning chains) rather than redistributing original video content, ensuring compliance with copyright and licensing requirements. Annotators were informed that their labels would be used for academic research and model training, with no personally identifiable information collected or retained.

\begin{table}[t]
  \centering
\resizebox{\columnwidth}{!}{%
\begin{tabular}{lccccc}
    \toprule
\textbf{Dataset} & \textbf{Domain} & \textbf{Video} & \textbf{Audio} & \textbf{Narrative} \\
    \midrule
QVHighlights & Vlog/News & \ding{51} & \ding{55} & \ding{55} \\
Charades-STA & Activity & \ding{51} & \ding{55} & \ding{55} \\
TVR & TV Show & \ding{51} & \ding{55} & \ding{55} \\
ShortVID & Short Video & \ding{51} & \ding{51} & \ding{55} \\
MomentSeeker & Multi-domain & \ding{51} & \ding{55} & \ding{51} \\
\midrule
\textbf{StoryTR (Ours)} & \textbf{Short Drama} & \textbf{\ding{51}} & \textbf{\ding{51}} & \textbf{\ding{51}} \\
    \bottomrule
  \end{tabular}%
}
\caption{Comparison with existing VMR benchmarks. StoryTR is the only dataset combining full modality (video + audio) with narrative-focused queries requiring character intent and plot reasoning.}
\label{tab:dataset_comparison}
\end{table}
\subsection{Query Types: A Cognitive Hierarchy}

StoryTR queries are organized into three tiers of increasing cognitive complexity, designed to systematically probe the perception-cognition gap:

\begin{itemize}[leftmargin=*,itemsep=2pt,topsep=2pt]
\item \textbf{Tier 1: Intent Decoding} (First-order ToM). These queries require inferring the underlying purpose or motivation behind character actions beyond surface-level observations. Models must reason about \textit{why} characters behave as they do, not just \textit{what} they do. Example: ``Find the explanation for why Charles is not present.'' Success requires understanding that absence implies a causal reason that must be explicitly stated or implied in dialogue or narration.

\item \textbf{Tier 2: Narrative Reasoning} (Second-order ToM). These queries demand understanding story structure, plot progression, and causal logic across temporal sequences. Models must trace how narrative states evolve and identify pivotal moments where relationships, knowledge states, or power dynamics shift. Example: ``Locate the moment when their relationship shifts from conflict to reconciliation.'' This requires recognizing abstract narrative transitions rather than concrete visual events.

\item \textbf{Tier 3: Boundary Localization} (Evidence Grounding). These queries require precise temporal grounding of abstract narrative concepts to specific multimodal evidence. Models must identify the exact boundaries where implicit mental states become manifest through observable cues (dialogue, expressions, actions, music). Example: ``Find evidence demonstrating the character's growing suspicion.'' Success requires aggregating subtle cues across modalities to pinpoint when abstract emotions crystallize into detectable signals.
\end{itemize}

The detailed reasoning process for generating these queries is described in Section~\ref{sec:selfqa_agent}.

\begin{figure*}[t]
\centering
\includegraphics[width=0.95\textwidth]{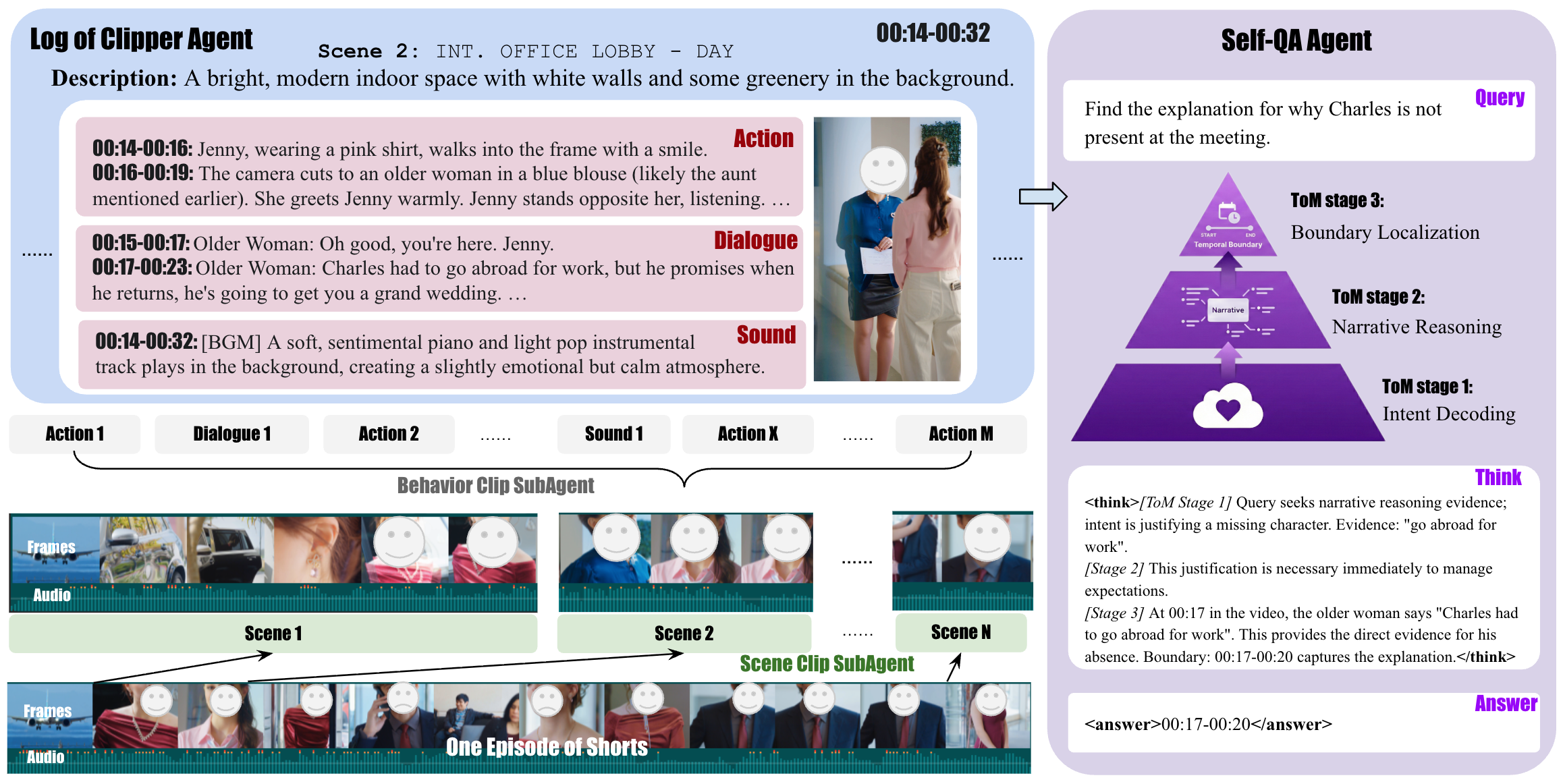}
\caption{Overview of our native multimodal perception pipeline for narrative shorts (short dramas/reels). We leverage complementary foundation models: Gemini-3.0-Pro for high-quality annotation and ARC-Hunyuan for efficient feature extraction and training. The pipeline processes short drama videos through multimodal encoding to generate temporal localization results with detailed reasoning logs.}
\label{fig:overview}
\end{figure*}
\subsection{Comparison with Existing Benchmarks}

Table~\ref{tab:dataset_comparison} positions StoryTR within the VMR landscape. The key distinction is not domain (short drama vs. vlog) but \textit{cognitive requirement}. StoryTR is the only benchmark requiring Theory of Mind reasoning. Its queries probe character intent, emotional dynamics, and narrative causality rather than observable actions. Combined with native multimodal input (video + audio), StoryTR provides the first rigorous testbed for evaluating whether models can bridge the perception-cognition gap.

\section{Methodology}


Our methodology is grounded in a key insight: \textbf{ToM reasoning is learnable through data, not architecture}. Traditional VMR annotations provide only timestamps. They tell models \textit{where} to look but not \textit{why} a moment matters. This supervision gap explains why scaling model parameters alone fails to bridge the perception-cognition divide.

We address this through a \textbf{ToM-Guided Data Paradigm}: generating training data that contains explicit reasoning chains, enabling smaller models to acquire cognitive capabilities through chain-of-thought supervision. As illustrated in Figure~\ref{fig:overview}, we implement this paradigm via an Agentic Data Pipeline with two components: a \textit{Clipper Agent} that transforms raw video into structured perceptual logs, and a \textit{Self-QA Agent} that synthesizes query-answer pairs with explicit three-tier ToM reasoning. The pipeline distills Gemini-3.0-Pro's narrative understanding into the open-source ARC-Hunyuan model.

\subsection{Clipper Agent: Perception Foundation}

The Clipper Agent transforms raw video into structured perceptual logs. These logs serve as the \textit{sensory evidence} upon which ToM reasoning operates. This addresses a key challenge: ToM requires reasoning about cues that may exist only in non-visual modalities (e.g., a dissonant chord signaling hidden tension in a visually calm scene).

\textbf{Multimodal Log Generation.} Driven by Gemini-3.0-Pro's native multimodal capabilities, the agent produces three temporally-aligned streams. First, \textit{Actions} capture fine-grained physical behaviors (``woman raises hand to face and wipes tears'') with character identification via physical descriptors to avoid hallucination. Second, \textit{Dialogue} records transcribed speech with speaker attribution. Third, \textit{Sounds} include both diegetic audio (footsteps, door slams) and non-diegetic elements (BGM shifts). The third stream is critical for short dramas, which heavily rely on music to signal emotional tone invisible to visual analysis.

\textbf{Shot-Aware Tracking.} A key design principle is explicit re-identification across camera cuts. This prevents the common error of merging actions from different characters across shot boundaries. It ensures the logs provide accurate evidence for downstream reasoning.

The output is a dense, timestamped ``screenplay'' (Figure~\ref{fig:overview}) that bridges raw multimodal signals to symbolic representations suitable for cognitive reasoning. Importantly, this perception layer is an \textit{enabler}, not the contribution itself. It provides the evidence base upon which ToM reasoning operates.

\subsection{Self-QA Agent: ToM Reasoning Engine}
\label{sec:selfqa_agent}

The Self-QA Agent is the core of our data paradigm. While the Clipper Agent captures \textit{what} happens, the Self-QA Agent generates training data that teaches models \textit{why} moments matter. This is the cognitive reasoning missing from traditional VMR annotations.

\textbf{Three-Tier ToM Reasoning.} The agent produces query-answer pairs with explicit reasoning chains that mirror human narrative understanding:

\textit{Tier 1: Intent Decoding} (First-order ToM). Given a query like ``Find the explanation for why Charles is not present,'' the agent identifies that this seeks narrative reasoning, not visual detection. The underlying intent is to justify a character's absence; key evidence markers include dialogue containing explanatory phrases. This tier teaches models to distinguish queries requiring cognitive reasoning from those requiring perception.

\textit{Tier 2: Narrative Reasoning} (Second-order ToM). The agent anchors the query within story structure, reasoning that absence explanations typically appear early in a scene to manage audience expectations. This tier generates explicit chains explaining \textit{why} a specific segment constitutes the correct match. It teaches models the causal logic of narrative construction.

\textit{Tier 3: Boundary Localization} (Evidence Grounding). The agent grounds abstract concepts in concrete evidence: ``At 00:17, the older woman states `Charles had to go abroad for work,' providing direct evidence.'' The boundary 00:17--00:20 captures the complete dialogue turn. This tier ensures models learn to retrieve segments grounded in explicit multimodal evidence while respecting narrative arc completeness.

\textbf{Data as Curriculum.} The resulting training data functions as a cognitive curriculum. Rather than learning implicit patterns from timestamps alone, models acquire explicit reasoning strategies through chain-of-thought supervision. This paradigm enables knowledge transfer by distilling Gemini-3.0-Pro's ToM capabilities into smaller, deployable models.

\subsection{Knowledge Distillation via Supervised Fine-Tuning}

\textbf{Backbone Selection.} We select ARC-Hunyuan~\cite{ge2025archunyuan} (7B parameters) as the student model. It is a unified video-language model with native audio processing via Whisper integration. The choice is deliberate: we aim to demonstrate that \textit{reasoning capability can be transferred through data}, independent of model scale.

\textbf{ToM Distillation.} We perform supervised fine-tuning using the Self-QA Agent's outputs: query-answer pairs enriched with three-tier reasoning chains. This process distills Gemini-3.0-Pro's ToM capabilities into the 7B model, teaching it not just \textit{where} to localize but \textit{how} to reason about narrative structure. The resulting model, \textbf{Shorts-Moment}, acquires cognitive capabilities that perception-only training cannot provide.

\begin{table*}[t]
\centering
\resizebox{\textwidth}{!}{%
\begin{tabular}{l|cccc|cccc|cccc}
\toprule
\multirow{2}{*}{Models} & \multicolumn{4}{c|}{Precision} & \multicolumn{4}{c|}{Recall} & \multicolumn{4}{c}{IoU} \\
\cmidrule(lr){2-5} \cmidrule(lr){6-9} \cmidrule(lr){10-13}
 & @0.3 & @0.5 & @0.8 & AVG & @0.3 & @0.5 & @0.8 & AVG & @0.3 & @0.5 & @0.8 & AVG \\
\midrule
\multicolumn{13}{l}{\textit{Closed-source Models}} \\
\midrule
\underline{Gemini-3.0-Pro} & 0.804 & 0.698 & 0.474 & 0.659 & 0.866 & 0.795 & 0.551 & 0.718 & 0.757 & 0.580 & 0.247 & 0.532 \\
Gemini-2.5-Pro & 0.766 & 0.665 & 0.462 & 0.630 & 0.840 & 0.789 & 0.541 & 0.706 & 0.719 & 0.562 & 0.233 & 0.505 \\
Gemini-2.5-Flash & 0.394 & 0.365 & 0.276 & 0.344 & 0.369 & 0.310 & 0.181 & 0.293 & 0.331 & 0.245 & 0.085 & 0.231 \\
\midrule
\multicolumn{13}{l}{\textit{Open-source Baselines}} \\
\midrule
Qwen3-Omni & 0.152 & 0.127 & 0.089 & 0.130 & 0.127 & 0.098 & 0.047 & 0.099 & 0.110 & 0.065 & 0.013 & 0.074 \\
ARC-Hunyuan\cite{ge2025archunyuan} & 0.551 & 0.457 & 0.281 & 0.447 & 0.647 & 0.599 & 0.406 & 0.542 & 0.487 & 0.353 & 0.127 & 0.344 \\
\midrule
\multicolumn{13}{l}{\textit{Ours}} \\
\midrule
\multirow{2}{*}{Shorts-Moment (Ours)} & \textbf{0.631} & \textbf{0.521} & \textbf{0.309} & \textbf{0.493} & \textbf{0.698} & \textbf{0.622} & \textbf{0.420} & \textbf{0.570} & \textbf{0.572} & \textbf{0.420} & \textbf{0.158} & \textbf{0.396} \\
 & \scriptsize{(+14.5\%)} & \scriptsize{(+14.0\%)} & \scriptsize{(+10.0\%)} & \scriptsize{(+10.3\%)} & \scriptsize{(+7.9\%)} & \scriptsize{(+3.8\%)} & \scriptsize{(+3.4\%)} & \scriptsize{(+5.2\%)} & \scriptsize{(+17.5\%)} & \scriptsize{(+19.0\%)} & \scriptsize{(+24.4\%)} & \scriptsize{(+15.1\%)} \\
\bottomrule
\end{tabular}%
}
\caption{Video Moment Retrieval Evaluation Results. We report accuracy at different thresholds (@0.3, @0.5, @0.8) and the mean (AVG). For our SFT model (Shorts-Moment), we additionally report the relative improvement over the baseline (ARC-Hunyuan) in parentheses. Best results in each category are marked in \textbf{bold}. We \underline{underline} Gemini-3.0-Pro to denote its role as the teacher model for data generation, which also achieves the best performance among closed-source models.}
\label{tab:evaluation_results}
\end{table*}
\section{Experiments}


Our experiments test three hypotheses derived from our theoretical framework:
\begin{itemize}[leftmargin=*,itemsep=0pt,topsep=2pt]
\item \textbf{H1 (Perception-Cognition Gap)}: Current models, despite strong perceptual capabilities, fail on narrative retrieval. This validates that the gap exists.
\item \textbf{H2 (ToM Learnability)}: Explicit ToM reasoning chains improve temporal precision. This validates that cognitive capabilities can be taught through data.
\item \textbf{H3 (Reasoning $>$ Scale)}: A 7B model with ToM training outperforms larger models without it. This validates that cognitive depth matters more than parameter count.
\end{itemize}

\subsection{Experimental Setup}

We evaluate on the StoryTR test set (811 samples). Performance is measured using Intersection over Union (IoU) at thresholds $\theta \in \{0.3, 0.5, 0.8\}$, along with Precision and Recall.

\textbf{Video and Audio Sampling.} Following ARC-Hunyuan's protocol, we apply adaptive sampling based on video duration. For videos $\leq$150s, we sample at 1 FPS (one frame per second); for longer videos, we uniformly sample 150 frames. Each frame is selected from the midpoint of its corresponding time segment. For audio, we use 16kHz sampling rate with a maximum of 150 segments. Each segment captures 2 seconds of audio; for videos exceeding 300s, we uniformly sample 150 segments across the timeline. Audio shorter than 1 second is zero-padded.

\textbf{Input Format.} Following ARC-Hunyuan\cite{ge2025archunyuan}'s protocol, we construct inputs by concatenating $N$ visual frame tokens with the text query, where each frame token encodes one sampled image. Audio features are extracted via the Whisper encoder and fused with visual tokens before being fed to the model.

\textbf{Baselines.} We compare against both closed-source and open-source models (see Appendix~\ref{app:baseline} for detailed specifications). For closed-source baselines, we evaluate the Gemini model family: Gemini-3.0-Pro serves as the teacher model for our data generation pipeline, representing Google's most advanced native multimodal architecture with 1M context length; Gemini-2.5-Pro provides a high-performance comparison point with advanced post-training via reinforcement learning; Gemini-2.5-Flash serves as an efficiency baseline as a lightweight distilled variant. For open-source baselines, we evaluate Qwen3-Omni, a 30B Mixture-of-Experts model capable of end-to-end omni-modal processing, and ARC-Hunyuan, a dense 7B model with timestamp overlay mechanism designed for temporal localization, which serves as the backbone for our Shorts-Moment model.

\begin{figure*}[t]
\centering
\includegraphics[width=0.95\textwidth]{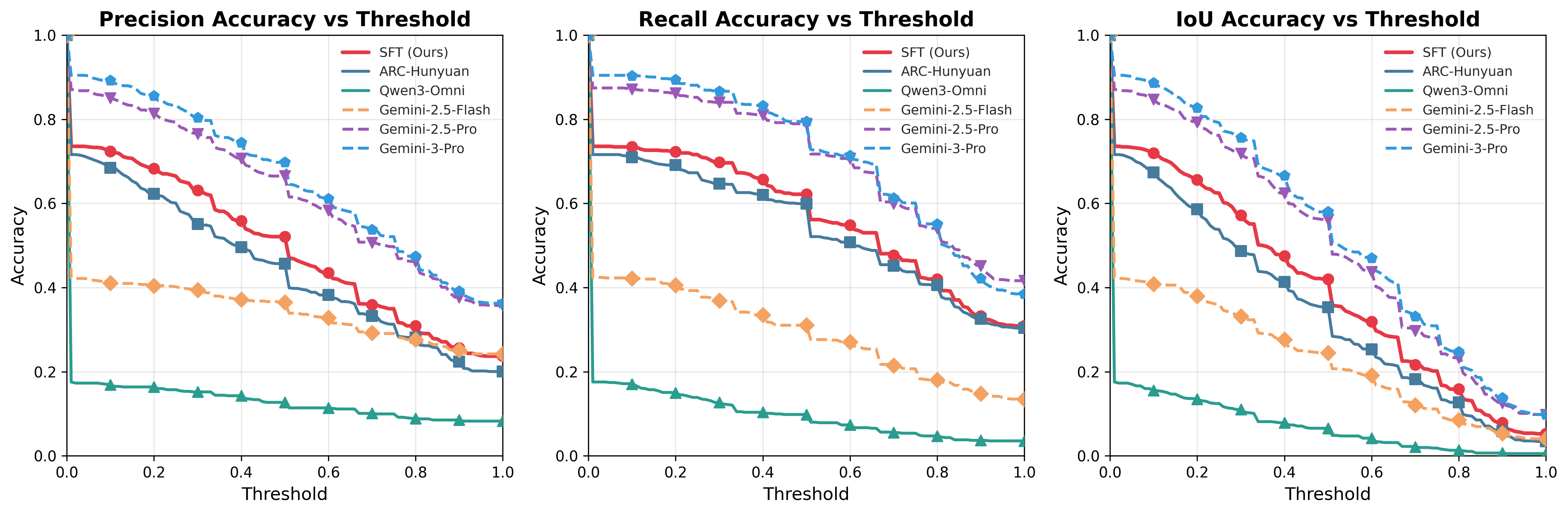}
\caption{Accuracy results for curves}
\label{fig:accuracy_curves}
\end{figure*}

\subsection{H1: The Perception-Cognition Gap Exists}

Table~\ref{tab:evaluation_results} quantifies the severity of the gap. Despite strong performance on action-centric benchmarks, models collapse on narrative retrieval:

\textbf{Perception Alone Fails.} Qwen3-Omni (30B MoE) achieves only 0.074 Avg IoU. Despite state-of-the-art perceptual capabilities, this model fails catastrophically when queries require reasoning about intent. Gemini-2.5-Flash, optimized for efficiency, drops to 0.231. Even Gemini-3.0-Pro, the strongest baseline, achieves only 0.532 Avg IoU.

\textbf{The Gap Is Cognitive, Not Perceptual.} These models can detect ``a woman smiling'' but cannot infer ``she is concealing hostility.'' StoryTR's Tier 2-3 queries (60\% of test set) require exactly this cognitive leap. This explains why perception-only models fail despite adequate sensory processing.

\subsection{H2: ToM Reasoning Is Learnable}

\textbf{Precision at Strict Thresholds.} The most striking result appears at IoU@0.8: our model improves +24.4\% over the baseline. This validates that ToM reasoning enables \textit{precise} boundary localization. The three-tier process (intent → narrative → evidence) helps models pinpoint exact narrative boundaries rather than approximate vicinities.

\textbf{Explicit Chains Outperform Implicit Learning.} Base ARC-Hunyuan (trained on timestamps only) achieves 0.344 Avg IoU. It identifies general locations but lacks precision. Our ToM-enhanced model achieves 0.396 (+15.1\%), demonstrating that explicit reasoning chains teach models \textit{why} boundaries matter, not just \textit{where} they are.

\subsection{H3: Reasoning Capability $>$ Parameter Scale}

\textbf{7B Beats 30B.} Our 7B Shorts-Moment outperforms the 30B Qwen3-Omni by 5$\times$ in Avg IoU (0.396 vs 0.074). This validates our core thesis: cognitive depth matters more than computational scale. A model \textit{taught to reason} outperforms a larger model that merely perceives.

\textbf{Approaching Closed-Source SOTA.} Our model surpasses Gemini-2.5-Flash (0.396 vs 0.231 Avg IoU) and narrows the gap to Gemini-3.0-Pro (0.396 vs 0.532). It achieves 74\% of the teacher's performance with a fraction of parameters. This validates that ToM capabilities can be effectively distilled through our data paradigm.

\section{Discussion and Analysis}

\begin{figure*}[t]
    \centering
    \includegraphics[width=0.95\textwidth]{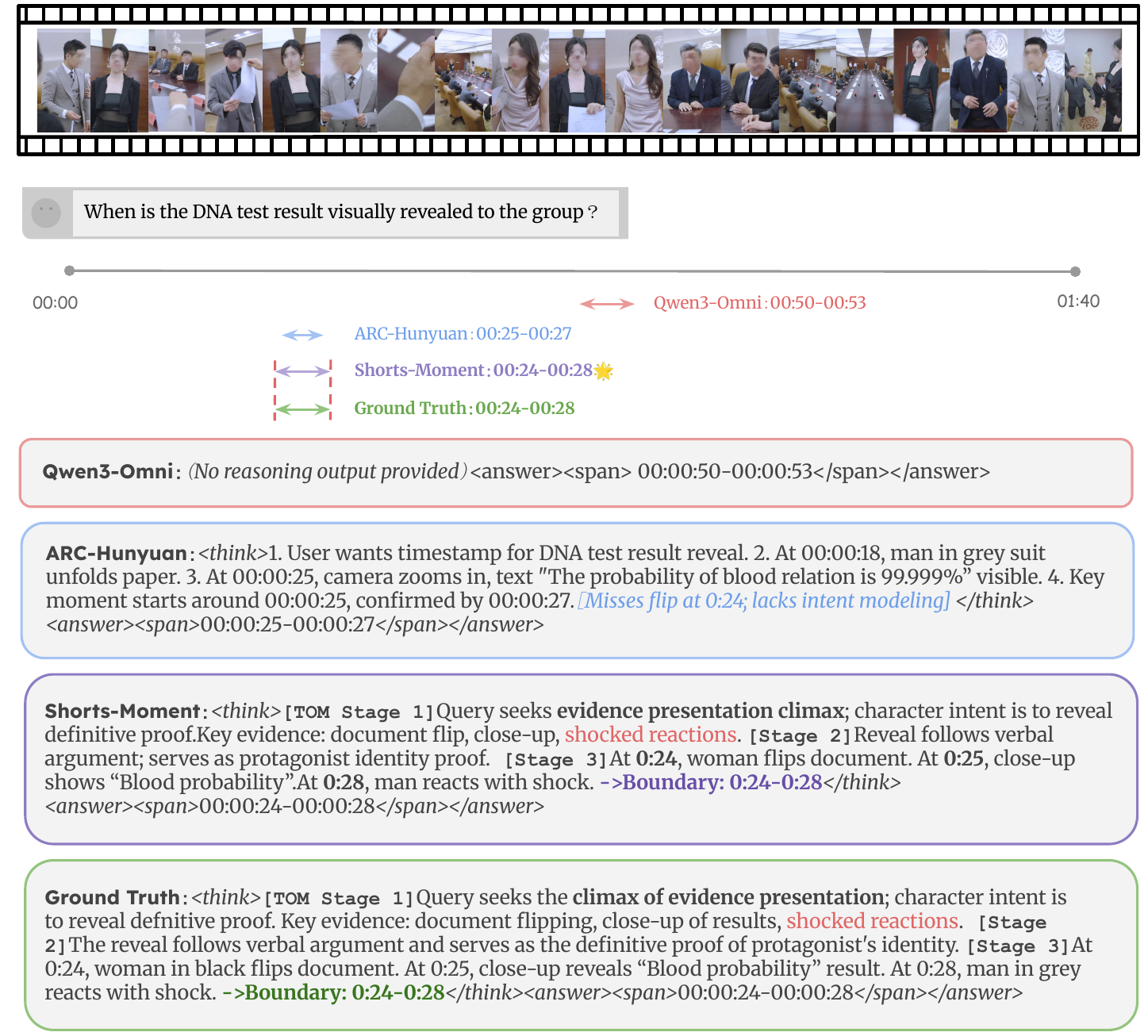}
    \caption{Case study comparing reasoning outputs for query ``\textit{When is the DNA test result visually revealed to the group?}''. Shorts-Moment achieves exact match (IoU=1.0) via three-stage ToM reasoning, while ARC-Hunyuan misses the boundary (IoU=0.5) and Qwen3-Omni fails completely (IoU=0.0).}
    \label{fig:case_study}
\end{figure*}

\subsection{Threshold Sensitivity Analysis}

Figure~\ref{fig:accuracy_curves} presents accuracy curves across varying IoU thresholds, revealing model robustness for temporal localization. As thresholds increase from 0.0 to 1.0, all models degrade, but at different rates.

\textbf{Performance Stability.} Our model demonstrates superior stability in the mid-range region (0.3-0.6). In Precision, we maintain >50\% accuracy until threshold 0.5, while ARC-Hunyuan drops faster. Similarly, Recall shows sustained higher rates across all thresholds, indicating improved coverage of narrative moments. It suggests that ToM-guided training teaches models to identify \textit{semantically coherent} boundaries rather than arbitrary cut points, since narrative beats naturally align with specific temporal spans.

\textbf{Temporal Localization Robustness.} Our model exhibits graceful degradation compared to open-source baselines. While Qwen3-Omni collapses at moderate thresholds, we maintain meaningful IoU scores up to 0.7 through narrative-aware fine-tuning that respects story boundaries. The divergence is especially pronounced in the IoU curve. The gap widens as thresholds increase, indicating that ToM reasoning provides the most value when \textit{precise} temporal grounding is required. This is exactly the regime where understanding ``why'' a moment matters becomes critical for determining ``where'' it begins and ends.

\textbf{Comparison with Closed-Source Models.} Gemini-2.5-Flash declines steeply after threshold 0.3, while Gemini-2.5-Pro and 3.0-Pro maintain stability at the cost of larger sizes. Our 7B model closely tracks these larger models in Recall while offering deployment advantages. Notably, we intersect with Gemini-2.5-Flash at threshold 0.4 in Precision and maintain superiority thereafter. This demonstrates that specialized training enables better discrimination at stricter requirements, where narrative understanding matters most.

\subsection{Case Study}

Figure~\ref{fig:case_study} crystallizes the perception-cognition gap. The query ``\textit{When is the DNA test result visually revealed to the group?}'' appears simple but encodes a critical semantic distinction. \textit{Revealed} implies \textbf{intentional disclosure}, a social act, not mere visual presence. The ground truth spans 00:24--00:28: the woman's deliberate flip (0:24), the document close-up (0:25), and the man's shocked reaction (0:28). These form a complete \textit{communicative arc}.

\textbf{Perception-Only Models Fail.} ARC-Hunyuan predicts 00:25--00:27 (IoU=0.5), starting one second late because it anchors on the camera zoom-in rather than the character's initiating action. Its reasoning, \textit{``camera zooms in showing the probability,''} reveals pure visual detection without intent modeling. It sees the document but misses the \textit{act of revealing}. Qwen3-Omni fails catastrophically (IoU=0.0), predicting 00:50--00:53. This 26-second offset demonstrates that parameter scale cannot substitute for cognitive capability.

\textbf{ToM Reasoning Succeeds.} Shorts-Moment achieves exact match (IoU=1.0) by reasoning through all three tiers. It identifies that the reveal \textit{begins} at 0:24 when the woman flips the document. It understands this initiating action is semantically part of ``revealing,'' not preparation. This requires \textit{intent modeling}: the flip is deliberate disclosure, not incidental handling. The model extends to 0:28 to capture the shocked reaction, completing the narrative beat.

\textbf{Generalization.} This pattern generalizes: queries involving social verbs (\textit{reveal, confront, comfort, deceive}) encode ToM concepts invisible to perception. ToM reasoning transforms temporal localization from ``when is X visible?'' to ``when does X achieve its \textit{narrative purpose}?''

\section{Conclusion}

This work addresses a fundamental gap in video understanding: current models can perceive \textit{what is happening} but fail to reason \textit{why it matters}. \textbf{StoryTR} provides the first rigorous testbed for narrative video moment retrieval, with queries explicitly designed to probe ToM capabilities across three cognitive tiers. Further, our \textbf{ToM-Guided Data Paradigm} shows that explicit reasoning chains can transfer cognitive capabilities from large to small models. It addresses the ``invisible intent'' problem through chain-of-thought supervision rather than parameter scaling. Our experiments validate a key insight for the field: \textbf{reasoning capability matters more than scale}. 

\section*{Limitations}

\textbf{ToM Complexity.} Our three-tier ToM framework captures first- and second-order reasoning but may not fully address higher-order ToM (e.g., ``A believes that B believes that C intends...''). Extending to deeper recursive reasoning remains future work.

\textbf{Cultural Variation.} ToM reasoning is culturally situated. Narrative conventions differ across traditions. Our dataset (55\% Chinese, 45\% English) may not generalize to all storytelling cultures without adaptation.

\textbf{Teacher Dependency.} The data paradigm relies on Gemini-3.0-Pro for reasoning chain generation. While we demonstrate successful distillation to 7B models, the initial data creation requires access to frontier models.

\textbf{Benchmark Scale.} StoryTR contains 8.1k samples. This is sufficient to validate our hypotheses but smaller than million-scale datasets. Scaling while maintaining annotation depth is an open challenge.

\section*{Ethics Statement}

This research involves video content analysis and human annotation, raising several ethical considerations that we have carefully addressed:

\textbf{Data Collection and Consent.} All videos used in StoryTR are legally obtained from publicly available platforms with proper licensing for research use. We secured explicit permission from content distributors for academic research purposes. We do not distribute copyrighted video content—only metadata annotations (queries, timestamps, reasoning chains)—ensuring compliance with copyright and fair use principles.

\textbf{Privacy.} Our dataset contains only commercially released content with no private or sensitive personal information. All content creators are professional actors performing in scripted productions. No surveillance footage, private recordings, or non-consensual content is included.

\textbf{Annotator Welfare.} Human annotation was conducted by trained researchers as part of their academic responsibilities. No crowdworkers were employed. Annotators were fully informed about the research purpose and provided consent for their annotations to be used in published datasets.

\textbf{Potential Misuse.} While our technology enables better video understanding for applications such as content retrieval and accessibility, it could potentially be misused for unauthorized content analysis or surveillance. We advocate for responsible use within legal and ethical boundaries, and recommend that future applications implement appropriate consent and transparency mechanisms.

\textbf{Bias and Representation.} Our dataset may reflect biases present in commercial short drama content, including potential under-representation of certain demographics, cultures, or narrative styles. Future work should address representation across diverse narratives, cultures, and production contexts. We encourage researchers using StoryTR to consider these limitations when deploying models in real-world applications.

\textbf{Dual Use.} We acknowledge that improved video understanding capabilities could have dual-use implications. We commit to responsible disclosure and encourage the community to develop appropriate safeguards for deployment.

\bibliography{custom}
\clearpage

\appendix
\section{Manual Annotation for Ground Truth Verification}
\label{sec:appendix_annotation}

\subsection{Annotation Setup}

To establish ground truth labels for evaluating our framework, we randomly selected 811 query-answer pairs (approximately 10\% of StoryTR) stratified across Chinese and English short dramas. The annotation was conducted by the authors using Label Studio \footnote{https://github.com/HumanSignal/label-studio}, an open-source platform

\begin{figure*}[t]
\centering
\includegraphics[width=0.95\textwidth]{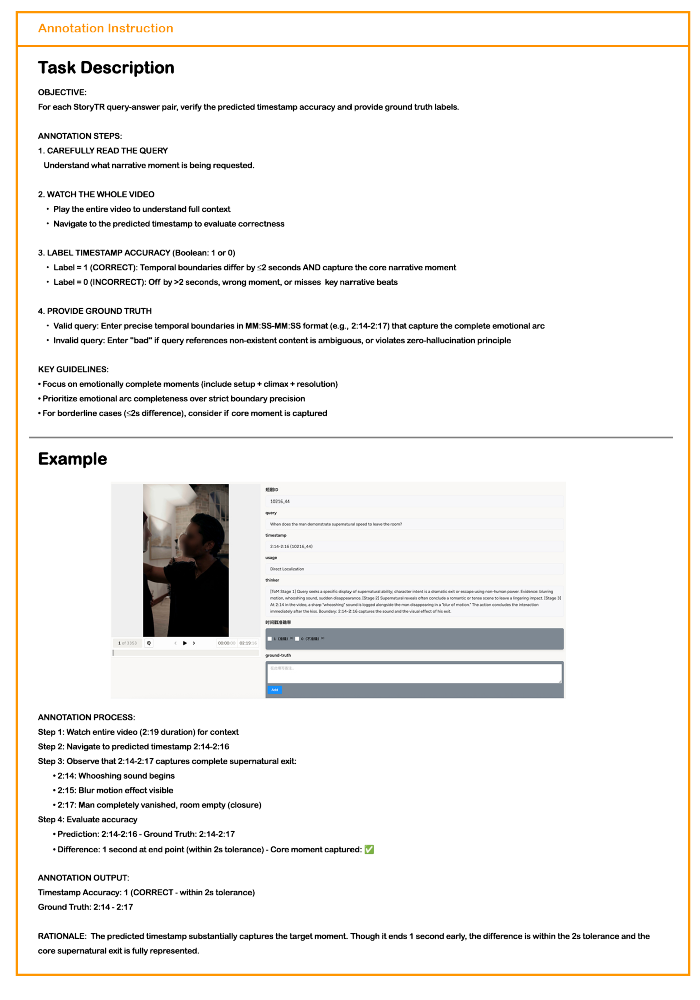}
\caption{Annotation interface and task description. \textbf{Top}: Complete task description including annotation steps, guidelines.\textbf{Bottom}: Label Studio interface showing video player (left) and StoryTR entry fields (right) with annotation inputs, and a worked example demonstrating the labeling process for a supernatural speed query. }
\label{fig:annotation_interface}
\end{figure*}

As shown in Figure~\ref{fig:annotation_interface}, the interface displays the video on the left and StoryTR entry data on the right, including: \texttt{video\_id} (drama + episode identifier), \texttt{query} (question), \texttt{timestamp} (model prediction), \texttt{usage} (query type), and \texttt{think} (model reasoning). Annotators completed two fields: (1) \textbf{Timestamp Accuracy} (binary: 1 if prediction is within 2 seconds of ideal boundaries; 0 otherwise), and (2) \textbf{Ground Truth} (precise temporal boundaries in MM:SS-MM:SS format, or \texttt{bad} for invalid queries).

\begin{figure*}[t]
    \centering
    \noindent\fbox{%
    \begin{minipage}{0.96\textwidth}
        \ttfamily\small
        \setlength{\parskip}{0.5em}

        You are a video time positioning assistant. You need to locate the corresponding time segment in the video based on the given query.

        Video total duration: \{self.duration\}

        Query: \{self.query\}

        Please follow these rules when responding: \\
        1. Analyze the query within the \textless think\textgreater\ tag to understand the specific information to locate. \\
        2. Provide the precise time segment within the \textless answer\textgreater\ tag, marking timestamps with \textless span\textgreater\ tags. \\
        3. Format the output inside \textless answer\textgreater\ tags, converting all time ranges to the pattern \textless span\textgreater HH:MM:SS - HH:MM:SS\textless /span\textgreater. \\
        4. If the query content does not exist in the video, clearly state this.

        Generate the response:
    \end{minipage}%
    }
    \caption{\textbf{System Prompt for Video Moment Retrieval.} The figure illustrates the structured instructions provided to the model for temporal grounding tasks. The prompt enforces a specific output format using XML-style tags for easy parsing. Variables enclosed in curly braces (e.g., \{self.query\}) are dynamically replaced during inference.}
    \label{fig:system_prompt}
\end{figure*}

\subsection{Task Description and Annotation Protocol}

Annotators followed a standardized protocol: (1) read query and understand target moment, (2) watch video and navigate to predicted timestamp, (3) label accuracy (1 if within $\leq$ 2s tolerance; 0 otherwise), and (4) provide ground truth (precise MM:SS-MM:SS boundaries for valid queries; \texttt{bad} for invalid queries). Figure~\ref{fig:annotation_interface} presents detailed task instructions and a complete annotation example demonstrating the process.

\begin{table}[h]
\centering
\small
\begin{tabular}{lcc}
\toprule
\textbf{Category} & \textbf{Samples} & \textbf{Accuracy (\%)} \\
\midrule
\textbf{Overall} & 811 & 92.11 \\
\midrule
\multicolumn{3}{l}{\textit{By Language}} \\
Chinese & 480 & 90.62 \\
English & 311 & 94.26 \\
\midrule
\multicolumn{3}{l}{\textit{By Query Type}} \\
Direct Localization & 276 & 90.94 \\
Event Identification & 273 & 91.21 \\
Evidence Localization & 265 & 94.27 \\
\bottomrule
\end{tabular}
\caption{Timestamp accuracy across categories. Accuracy is defined as predictions within 2 seconds of ground truth boundaries.}
\label{tab:annotation_results}
\end{table}
\subsection{Annotator Recruitment and Compensation}

\textbf{Recruitment.} Manual annotation was conducted by the authors, who are graduate students and researchers with expertise in video understanding and natural language processing. All annotators are fluent in both English and Chinese, ensuring accurate evaluation of bilingual content in StoryTR.

\textbf{Training and Qualification.} Before formal annotation, annotators completed a training session using 50 practice samples to establish inter-annotator agreement on boundary definitions and quality criteria. Disagreements were resolved through discussion to ensure consistent annotation standards.

\textbf{Compensation.} As the annotation was performed by the research team as part of their academic responsibilities, no additional monetary compensation was provided beyond standard research assistantship stipends. The annotation workload (811 samples) was distributed across the team and completed over approximately 40 person-hours. This approach ensured high-quality annotations from domain experts while avoiding potential quality issues associated with crowdsourced annotation for tasks requiring deep narrative understanding.

\textbf{Demographic Context.} All annotators are based in research institutions in China, where the standard compensation for research assistants is commensurate with local cost of living and academic norms.

\subsection{Results and Analysis}

Table~\ref{tab:annotation_results} presents accuracy results. Overall accuracy is \textbf{92.11\%}, with Chinese short dramas achieving \textbf{90.62\%} and English achieving \textbf{94.26\%}. By query type, Direct Localization achieves \textbf{90.94\%}, Event Identification \textbf{91.21\%}, and Evidence Localization \textbf{94.27\%}.

\begin{table*}[t]
    \centering
    \small
    \renewcommand{\arraystretch}{1.1}
    \setlength{\tabcolsep}{8pt}
    \resizebox{\textwidth}{!}{%
    \begin{tabular}{l l l l c l}
    \toprule
    \textbf{Model} & \textbf{API/Version} & \textbf{Developer} & \textbf{Release} & \textbf{Context} & \textbf{Architecture} \\
    \midrule
    Gemini-3.0-Pro & gemini-3.0-pro-preview & Google & Nov 2025 & 1M & Native multimodal; SOTA \\
    Gemini-2.5-Pro & gemini-2.5-pro & Google & Mar 2025 & 1M & Native multimodal \\
    Gemini-2.5-Flash & gemini-2.5-flash & Google & Apr 2025 & 1M & Distilled model \\
    \midrule
    Qwen3-Omni & Qwen3-Omni-30B-A3B-Instruct & Alibaba & Sep 2025 & 32K & MoE (30B total) \\
    ARC-Hunyuan & ARC-Hunyuan-Video-7B & Tencent & Jul 2025 & 20K & Dense 7B \\
    \midrule
    \textbf{Shorts-Moment (Ours)} & Fine-tuned from ARC-Hunyuan & - & - & 20K & Dense 7B + ToM \\
    \bottomrule
    \end{tabular}%
    }
    \caption{\textbf{Model Specifications.} API/Version column shows the exact model identifier used in experiments. Shorts-Moment is our model fine-tuned from ARC-Hunyuan on StoryTR training set.}
    \label{tab:model_specs}
\end{table*}
\section{Baseline Implementation Details}
\label{app:baseline}

\subsection{Experimental Setup}
To ensure fair comparison across diverse architectures, all baseline experiments were conducted under identical conditions. All models were evaluated in a zero-shot setting with temperature $\tau=1.0$ to standardize generation randomness. Video inputs were standardized at 2 FPS for all multimodal models. Identical prompt templates were used across all models, with minor adaptations only for specific chat formats (e.g., ChatML for Qwen). 

Evaluations were performed on NVIDIA A800 (80GB) GPUs. Closed-source models were accessed via Google Vertex AI. Open-source models were deployed using vLLM (ARC-Hunyuan) and HuggingFace Transformers (Qwen3-Omni) with KV cache optimization.

\subsection{Model Specifications}
We compare our approach against five state-of-the-art multimodal baselines. Detailed specifications are summarized in Table \ref{tab:model_specs}.

\noindent \textbf{Gemini Model Family} (accessed via Google Vertex AI):
\begin{itemize}
    \item \textbf{Gemini-3.0-Pro} (API: gemini-3.0-pro-preview): Google's most advanced multimodal model. It features native support for text, video, and audio. In our experiments, it serves as the \textit{Teacher model} for data generation.
    \item \textbf{Gemini-2.5-Pro} (API: gemini-2.5-pro): A high-performance model utilizing advanced post-training techniques including reinforcement learning.
    \item \textbf{Gemini-2.5-Flash} (API: gemini-2.5-flash): A lightweight model designed for efficiency, serving as an efficiency baseline.
\end{itemize}

\noindent \textbf{Open-Source Baselines:}
\begin{itemize}
    \item \textbf{Qwen3-Omni} (Version: Qwen3-Omni-30B-A3B-Instruct): An omni-modal MoE model capable of end-to-end processing of text, audio, and video.
    \item \textbf{ARC-Hunyuan} (Version: ARC-Hunyuan-Video-7B): A dense 7B model with timestamp overlay mechanism for temporal localization. Serves as the backbone for our Shorts-Moment model.
\end{itemize}

\noindent \textbf{Our Model:}
\begin{itemize}
    \item \textbf{Shorts-Moment}: Fine-tuned from ARC-Hunyuan-Video-7B on StoryTR training set. Acquires ToM reasoning and narrative moment retrieval capabilities through our Agentic Data Pipeline.
\end{itemize}

\subsection{Prompt Strategy}
\label{sec:prompt_strategy}

To mitigate the influence of prompt engineering on evaluation metrics and ensure a strictly fair comparison, we employed a unified prompt template across all models. As illustrated in Figure~\ref{fig:system_prompt}, the identical instruction was fed to both the closed-source Gemini family and the open-source baselines (Qwen and Hunyuan) without any model-specific optimization. The prompt explicitly constraints the output format, requiring models to predict precise start and end timestamps alongside a reasoning rationale. By freezing the input instruction, we ensure that the observed performance variances are solely attributable to the intrinsic video understanding and reasoning capabilities of the respective architectures.

\end{document}